\documentclass[journal]{IEEEtran}

\usepackage{hyperref}

\usepackage{array}
\newcolumntype{P}[1]{>{\centering\arraybackslash}p{#1}}

\ifCLASSINFOpdf
\else
   \usepackage[dvips]{graphicx}
\fi
\usepackage{url}

\usepackage{lmodern}
\usepackage[T1]{fontenc}

\usepackage{algorithm}
\usepackage{algorithmic}[1]
\usepackage{graphicx}
\usepackage{color}
\usepackage{subfig}
\usepackage{amsmath}
\usepackage{textcomp}
\DeclareMathOperator*{\argmin}{arg\,min}
\usepackage{wrapfig}

\begin{document}

\title{Ensemble Noise Simulation to Handle Uncertainty about Gradient-based Adversarial Attacks}

\author{Rehana Mahfuz, \IEEEmembership{Student Member, IEEE}, Rajeev Sahay, \IEEEmembership{Student Member, IEEE}, \\and Aly El Gamal, \IEEEmembership{Senior Member, IEEE}
\thanks{R. Mahfuz, R. Sahay, and A. El Gamal are with the Department of Electrical and Computer Engineering, Purdue University, West Lafayette, IN, USA. Email: \{rmahfuz, sahayr, elgamala\}@purdue.edu.}}

\maketitle

\begin{abstract}
Gradient-based adversarial attacks on neural networks can be crafted in a variety of ways by varying either how the attack algorithm relies on the gradient, the network architecture used for crafting the attack, or both. Most recent work has focused on defending classifiers in a case where there is no uncertainty about the attacker's behavior (i.e., the attacker is expected to generate a specific attack using a specific network architecture). However, if the attacker is not guaranteed to behave in a certain way, the literature lacks methods in devising a strategic defense. We fill this gap by simulating the attacker's noisy perturbation using a variety of attack algorithms based on gradients of various classifiers. We perform our analysis using a pre-processing Denoising Autoencoder (DAE) defense that is trained with the simulated noise. We demonstrate significant improvements in post-attack accuracy, using our proposed ensemble-trained defense, compared to a situation where no effort is made to handle uncertainty. 

\end{abstract}


\IEEEpeerreviewmaketitle

\section{Introduction}

\IEEEPARstart{A}{dversarial} attacks on neural networks pose a serious threat to safety-critical systems that rely on the high accuracies of these neural networks. The imperceptibility of additive evasion attacks makes it difficult to even detect their existence. Recent work has attempted to tackle this issue by designing defenses against such attacks, mostly focusing on a scenario where the assumption is that the attacker has significant knowledge of the victim classifier, and hence will design an attack to optimally destroy the accuracy of that particular classifier.  
However, there is no guarantee that the attacker will choose to do so. Furthermore, adversarial examples transfer across classifiers, and an adversary could take advantage of this property by crafting an attack based on a different classifier. The attacker would do this when having only partial knowledge about the victim classifier, or when attempting to confuse the defender on purpose. Alternatively, another scenario is that the attacker is limited in computational resources, and may be trying to attack multiple classifiers at once. This is why they would tailor the attack to only one classifier, and use that to attack all classifiers. Therefore, there is a need to evaluate the sensitivity of the defense to the data used to train it, especially in the architecture mismatch case (i.e., when the architecture of the target classifier differs from the architecture of the adversary's classifier), since traditional strategies may not suffice \cite{transferability}.

In a defense-blind scenario, we investigate strategies to combat adversarial attacks in the presence of uncertainty about the classifier's architecture whose gradients were used to generate the attacks. We also tackle the uncertainty caused by the variability of the attack algorithm by considering three established attacks. \textbf{We empirically demonstrate the effectiveness of using an ensemble of noise simulations corresponding to all possible attacks for training a pre-processing Denoising Autoencoder (DAE)}.

{\bf Related Work:} 
Many existing methods to combat adversarial attacks assume that such attacks are specifically crafted using the gradient of the victim model. Often, these methods are, without modification, employed to mitigate attacks generated using a disparate classifier's gradient with less success than in the case where the attack is based on the victim classifier's gradient \cite{space_of_transf_adv_ex}. 
For example, defenses based on Principal Components Analysis (PCA) \cite{pca}, autoencoder-based dimensionality reduction \cite{cascade}, \cite{spl}, and denoising autoencoders \cite{spl} suffer a severe degradation of performance in architecture mismatch settings.
Recent work has proposed training multiple DAEs (for filtering instead of dimensionality reduction) and randomly selecting one as a defense at test time \cite{magnet}. While this may be effective in confusing the attacker, these DAEs have only been tested against very mild attacks. 

The idea of natural noise simulation has been extensively investigated before the successful application of deep learning over the past decade. For example, in \cite{sigproc8}, a noise compensation strategy is studied for speech recognition to accommodate common scenarios where training and test noises can emanate from different environments. Also, in \cite{sigproc10}, an early example for a pre-processing denoising recurrent network that precedes a time delay neural network was introduced. Further, in \cite{sigproc7}, multiple noise simulations were performed for robust speaker identification. Recently, using the new deep learning tools for simulating and filtering noise has also been investigated, particularly for speech processing (see e.g., \cite{sigproc3, sigproc12}). Simulating adversarial noise and directly using it for training the final classifier (a.k.a. adversarial training \cite{adv-train}) has been investigated in multiple settings (see e.g., \cite{sigproc1, sigproc5, sigproc11}), including settings where adversarial noise is used to model worst case scenarios for natural noise (see e.g., \cite{sigproc13,sigproc14}).

To our knowledge, when adversarial noise is used to model an attacker's perturbation, there has been little work in studying the transferability of defenses, particularly those relying on simulating adversarial noise. In an attempt to discuss the transferability of a particular defense across multiple neural networks, \cite{AAA} emphasizes that using a DAE is superior to adversarial training because a DAE requires training only once but can mitigate attacks on multiple classifiers that are trained to perform the same task. Using adversarial training would require retraining each of those classifiers. Not requiring classifier retraining could offer significant advantages in terms of computational cost and enabling applications (see e.g., \cite{sigproc4}). Given that the autoencoder defense is a transferable defense, our work attempts to determine how to best train it, when there is uncertainty about the gradient used to generate the attack and about the attack algorithm. Further, the ultimate goal is not to achieve robustness for a particular classifier, but to achieve robustness for a particular task. In that light, \cite{break} proposes training multiple classifiers to perform the same task and choosing one or more of them at test time, since it is difficult for the adversary to generate attacks using gradients of multiple classifiers. Also, they add a small Gaussian noise to the trained weights of the classifiers. However, randomly selecting one of many trained networks to perform the classification task and/or using majority vote neither prevents the adversarial examples from transferring to each of those trained networks, nor does it address the problem of uncertainty about the attack algorithm used by the adversary. Our work addresses both of these problems. Specifically, it addresses uncertainty about both the attack algorithm and the network whose gradients are used to generate the attack.

\section{Procedure}
We use a DAE as a defense, as explained in Algorithm 1. To generate data for DAE training, we consider two sources of variation: the attack algorithm and the classifier architecture whose gradients are used. For the attack algorithms, we consider three possibilities: Carlini-Wagner (CW) \cite{cw}, which is known for robustness, DeepFool (DF) \cite{df}, which adds a very small perturbation, and Fast Gradient Sign (FGS) \cite{fgs}, which enjoys computational efficiency. The CW attack minimizes the sum of the perturbation and a scaled cost function $\tilde{J}(\cdot)$ to find the optimum value of a variable $w$
\begin{equation}\label{eq:cw}
    \text{minimize  } c\cdot \tilde{J}(x') + ||x'-x||_2^2,
\end{equation}
where $\tilde{J}(x')$ is a cost function - very similar to cross-entropy - that penalizes the sample classification as the true label $t$: 
\begin{equation}
    \tilde{J}(x') = \max\{Z(x')_t -\max_{i\neq t}Z(x')_i,0\},
\end{equation}
and $x'=\sigma(w)$ is given by the sigmoid function, $Z(\cdot)_i$ is the input to the $i^{th}$ neuron of the softmax layer. The constant $c$ is chosen to be the smallest value such that $\tilde{J}(x^*) = 0$, where $x^*$ is the solution. Gradient descent with multiple random starting points is used to find the solution. 

\begin{algorithm}[htb]
   \label{alg:defense}

   \textbf{Function: }$dae$\textunderscore$generator$ \newline 
   \textbf{Input:}
   $x_{train}$, $list$ \textunderscore
   $attack$\textunderscore$alg$, $list$\textunderscore$w$\textunderscore$ model$\newline
  \hspace*{4mm} $x_{adv} = []$\newline
  \hspace*{4mm} for $attack$\textunderscore$alg$ in $list$\textunderscore$attack$\textunderscore$alg:$\newline
  \hspace*{7mm} for $w_{model}$ in $list$\textunderscore$w_{model}:$\newline
  \hspace*{10mm} $x_{adv}.extend(attack$\textunderscore$alg(x_{train}, w_{model}))$\newline
  \hspace*{4mm} $\mathit{shuf}$\textunderscore$idx = rand(len([x_{train}, x_{adv}]))$\newline
  \hspace*{4mm} $w_{dae} = train$\textunderscore$dae($\newline
  \hspace*{4mm} $input=\mathit{shuffle}([x_{train}, x_{adv}], idx = \mathit{shuf}$\textunderscore$idx),$\newline
  \hspace*{3mm} $target=\mathit{shuffle}([x_{train}]*\frac{len(x_{adv})}{len(x_{train})}+1, idx\!=\!\mathit{shuf}$\textunderscore$idx))$\newline
  \hspace*{4mm} return $w_{dae}$\newline
  \textbf{Function: }$test$\textunderscore$time$\textunderscore$defense$ \newline
    \textbf{Input:} $x_{test}$\newline
  \hspace*{4mm} return $w_{dae}.fprop(x_{test})$\newline
  \caption{We first extend the list $x_{adv}$ to include data $x_{train}$ that is attacked using every attack algorithm $attack$\textunderscore$alg$ from the list $list$\textunderscore$attack$\textunderscore$alg$, using gradients of every network with weights $w_{model}$ from the list $list$\textunderscore$w$\textunderscore$model$. A random permutation of indices from 1 to the sum of the lengths of $x_{train}$ and $x_{adv}$, $shuf$\textunderscore$idx$, is then found. Next, a DAE is trained with a combination of $x_{train}$ and $x_{adv}$, to produce as target output the corresponding clean data from $x_{train}$, both of which are shuffled according to $shuf$\textunderscore$idx$, to maintain consistency in order. To use this DAE as a defense, the function $test$\textunderscore$time$\textunderscore$defense$ forward propagates the test data $x_{test}$ through the trained DAE with weights $w_{dae}$, to obtain the reconstructed approximation of clean data.}
  
 \end{algorithm}

The DF attack tries to find a perturbed sample $x_{i+1}$ from $x_{i}$ by iteratively approximating the classifiers to be linear, and finding the minimum $l_2$-norm perturbation that leads to misclassification. Let $x_{i}$ be the sample at the $i^{th}$ iteration, $x_0 = x$, $C^*(x_0)$ be the true label, and $f_k(x)$ be the identity indicator for whether sample $x$ is classified to belong to class $k$.
The perturbation added at iteration $i$ is:
\begin{equation}\label{eq:df_soln}
     \dfrac{|f^{'}_l|}{||w^{'}_l||_2^2}w^{'}_l,
\end{equation}
where
\begin{equation}\label{eq: df_soln1}
l = \argmin_{k \neq C^*(x_0)}\dfrac{|f^{'}_k|}{||w^{'}_k||_2^2},
\end{equation}
\begin{equation} \label{eq: df_soln2}
    w^{'}_k = \bigtriangledown f_k(x_i) - \bigtriangledown f_{C^*(x_0)}(x_i),
\end{equation}
and
\begin{equation} \label{eq: df_soln3}
    f^{'}_k = f_k(x_i) - f_{C^*(x_0)}(x_i).
\end{equation}
This continues till misclassification, or for a certain number of iterations. In our experiments, this number is 50.

The FGS attack adds a perturbation proportional to the sign of the gradient of the cost function $J(\theta, x, y)$ with respect to the input. This perturbation is $l_2$-bounded.
\begin{equation}\label{eq:fgs}
    \widetilde{x}(\epsilon) = x + \epsilon * \frac{ \bigtriangledown_x J(\theta,x,y)}{||\bigtriangledown_x J(\theta,x,y)||_2}.
\end{equation}
For the classifier architecture on which the attacks are based, the two architecture types considered are Fully Connected (FC) and Convolutional Neural Network (CNN). Only a CNN is considered for the CIFAR-10 classification task, as it is difficult to solve with an FC classifier.
\subsection{Experimental Setup}
The MNIST-Digit and CIFAR-10 datasets were used for classification of 28$\times$28 and 32$\times$32 pixel images, respectively, into one of 10 classes. 
For fully connected networks, we use the notation FC-$n_1$-$n_2$-...-$n_l$ to denote that layer i has $n_i$ neurons, for $i \in \{1,\cdots,l\}$. The MNIST victim FC network has architecture FC-784-100-100-10, achieving an accuracy of 98.11\%, and the MNIST adversary's FC network has architecture FC-784-200-100-100-10. Each layer has a Rectified Linear Unit (ReLU) activation, except the final layer which has softmax. The MNIST adversary's CNN architecture is shown in Table \ref{table:mnist_adv_cnn}. The MNIST victim CNN achieves an accuracy of 98.66\%, and has a similar architecture to the adversary's, but with only two convolutional layers, followed by a softmax layer. The CIFAR-10 victim CNN, shown in Table \ref{table:cifar_victim_cnn}, achieves an accuracy of 90.44\%. Note that ELU and BatchNorm refer to Exponential Linear Unit activation and Batch Normalization, respectively. Also, note that the notation (Conv 3x3x32, ELU, BatchNorm)x2 refers to using two consecutive sequences of a convolutional layer followed by ELU activation and Batch Normalization. 
To obtain variations of this architecture to train the CIFAR-10 defense, an extra (Conv 3x3x$z$, ELU, BatchNorm) sequence is added after the second, fourth, or sixth such sequence, where $z$ is the number of filters in the convolutional layer preceding the first added convolutional layer. These variations are used only to train the defense, and not to generate attacks to test the defense against. The adversary's architecture is obtained by adding such a sequence after the eighth such sequence.
Data augmentation was performed prior to feeding the CIFAR-10 data into the CNN, through rotations of up to 15 degrees, width/height shifts of up to 10\% of the original, and horizontal flips\footnote{Details about the denoising autoencoders and training details for all networks can be found in Appendix.}.

\begin{table}
\begin{tabular}{ p{8cm} }
\hline
 Conv 3x3x32, ReLU\\
 Conv 3x3x64, ReLU\\
 Max Pool 2x2\\
 Dropout (rate = 0.25)\\
 FC (128 neurons), ReLU\\
 Dropout (rate = 0.5)\\
 Softmax (10 classes)\\
 \hline
 \end{tabular}
 \caption{MNIST adversary's CNN architecture.}
 \label{table:mnist_adv_cnn}
 \end{table}

\begin{table}
 \begin{tabular}{ p{8cm} }
 \hline
 (Conv 3x3x32, ELU, BatchNorm)x2\\
 Max Pool 2x2, Dropout (rate = 0.2)\\
 (Conv 3x3x64, ELU, BatchNorm)x2\\
 Max Pool 2x2, Dropout (rate = 0.3)\\
 (Conv 3x3x128, ELU, BatchNorm)x2\\
 Max Pool 2x2, Dropout (rate = 0.4)\\
 (Conv 3x3x128, ELU, BatchNorm)x2\\
 Max Pool 2x2, Dropout (rate = 0.4)\\
 Softmax (10 classes)\\
 \hline
 \end{tabular}
 \caption{CIFAR-10 victim CNN architecture.}
 \label{table:cifar_victim_cnn}
 \end{table}

\subsection{Approach}

\begin{figure}[!tbp]
  \centering
  \subfloat{\includegraphics[width=9mm]{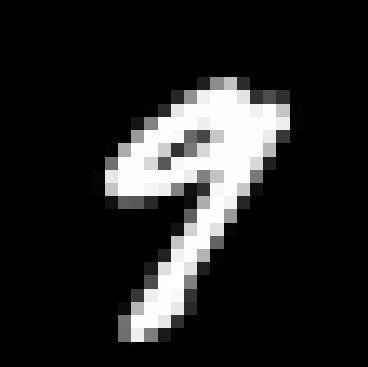}\label{fig:f1}}\hfill
  \subfloat{\includegraphics[width=9mm]{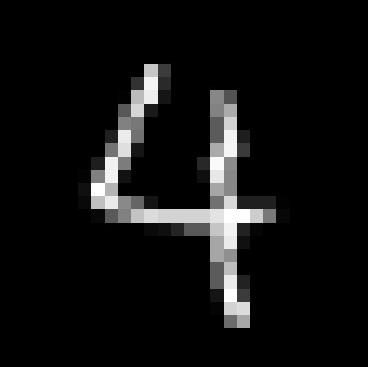}\label{fig:f1}}\hfill
  \subfloat{\includegraphics[width=9mm]{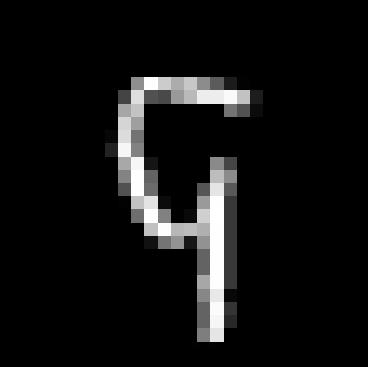}\label{fig:f1}}\hfill
  \subfloat{\includegraphics[width=9mm]{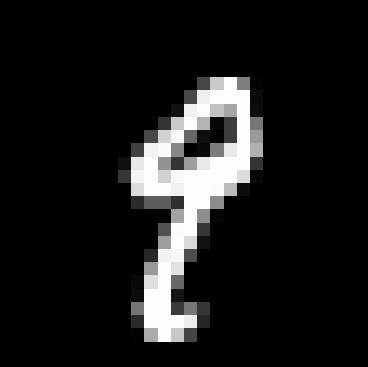}\label{fig:f1}}\hfill
  \subfloat{\includegraphics[width=9mm]{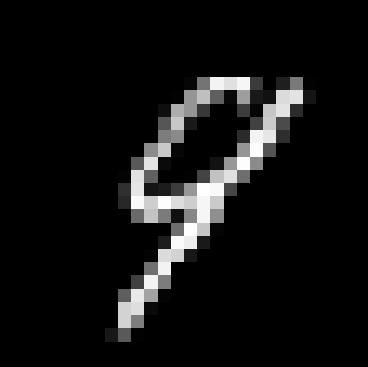}\label{fig:f1}}\hfill
  \subfloat{\includegraphics[width=9mm]{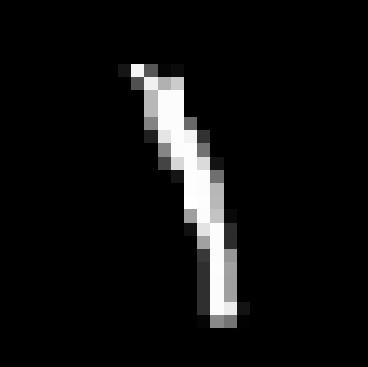}\label{fig:f1}}\hfill
  \subfloat{\includegraphics[width=9mm]{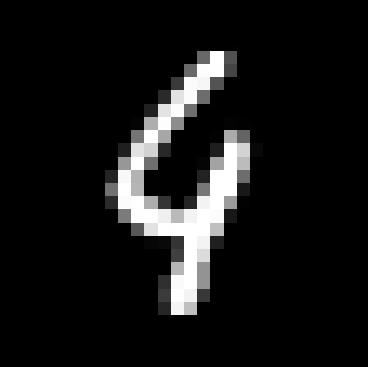}\label{fig:f2}}\hfill
  \subfloat{\includegraphics[width=9mm]{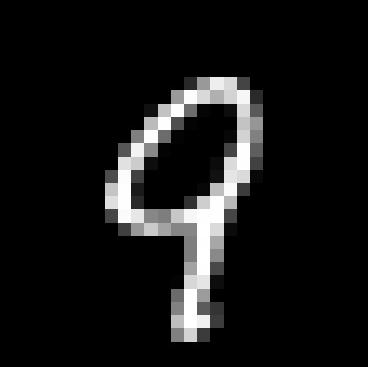}\label{fig:f2}}\hfill
  \subfloat{\includegraphics[width=9mm]{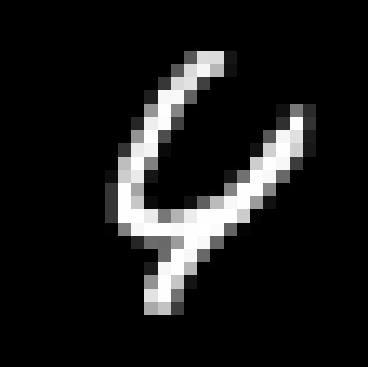}\label{fig:f2}}\hfill
  \subfloat{\includegraphics[width=9mm]{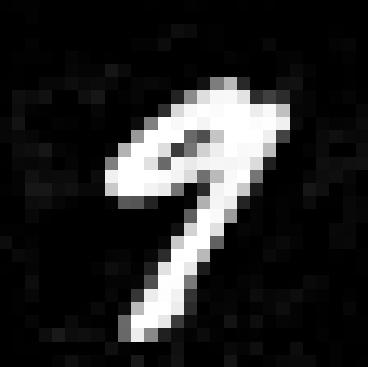}\label{fig:f1}}\hfill
  \subfloat{\includegraphics[width=9mm]{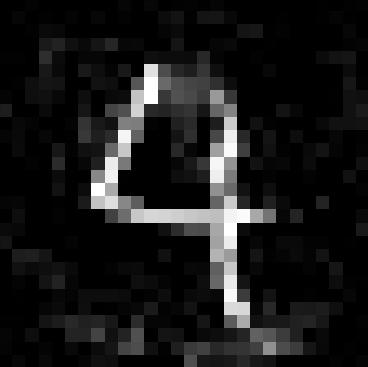}\label{fig:f1}}\hfill
  \subfloat{\includegraphics[width=9mm]{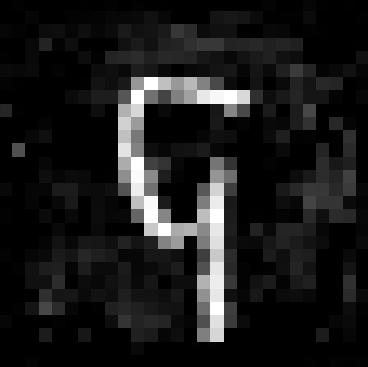}\label{fig:f1}}\hfill
  \subfloat{\includegraphics[width=9mm]{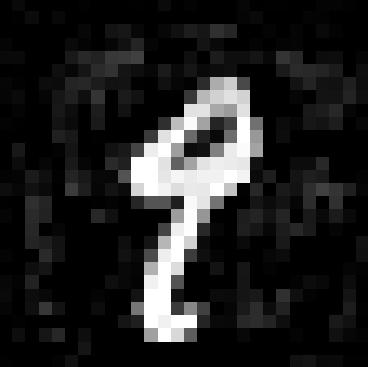}\label{fig:f1}}\hfill
  \subfloat{\includegraphics[width=9mm]{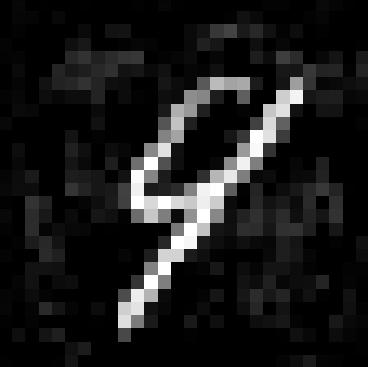}\label{fig:f1}}\hfill
  \subfloat{\includegraphics[width=9mm]{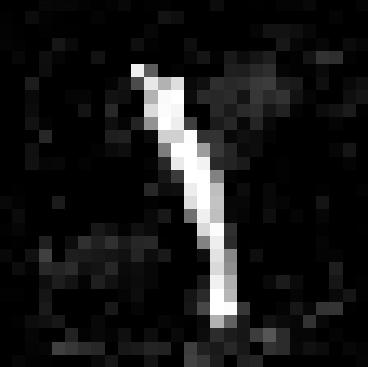}\label{fig:f1}}\hfill
  \subfloat{\includegraphics[width=9mm]{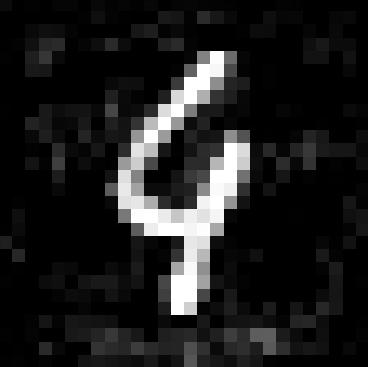}\label{fig:f2}}\hfill
  \subfloat{\includegraphics[width=9mm]{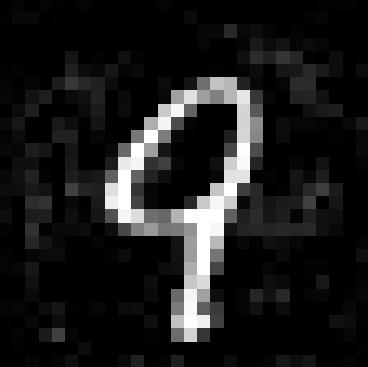}\label{fig:f2}}\hfill
  \subfloat{\includegraphics[width=9mm]{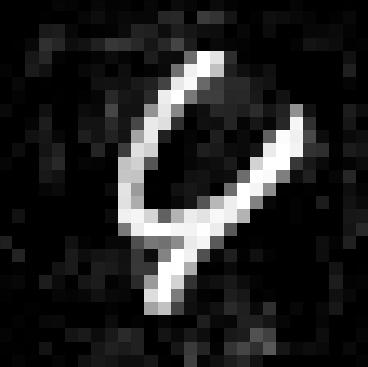}\label{fig:f2}}\hfill
  \subfloat{\includegraphics[width=9mm]{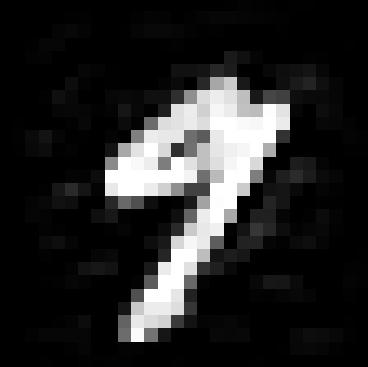}\label{fig:f1}}\hfill
  \subfloat{\includegraphics[width=9mm]{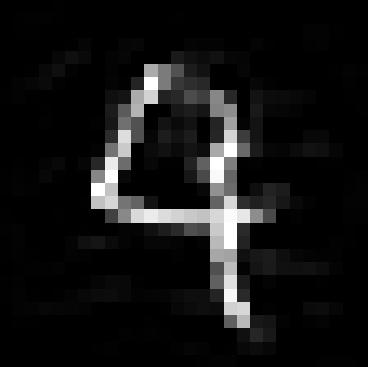}\label{fig:f1}}\hfill
  \subfloat{\includegraphics[width=9mm]{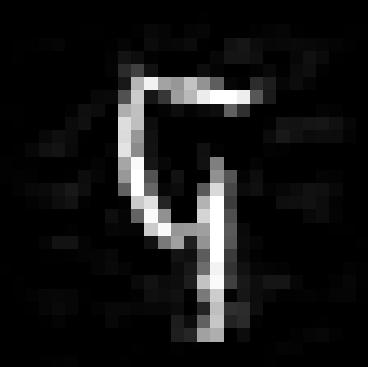}\label{fig:f1}}\hfill
  \subfloat{\includegraphics[width=9mm]{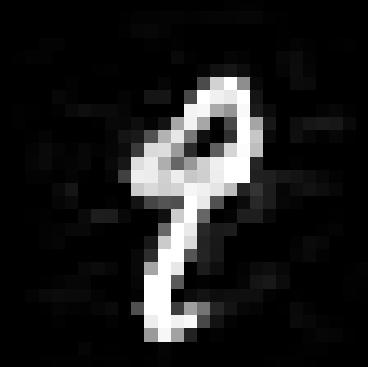}\label{fig:f1}}\hfill
  \subfloat{\includegraphics[width=9mm]{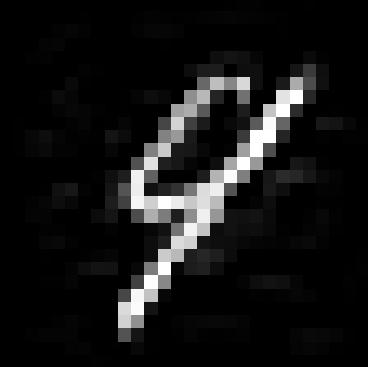}\label{fig:f1}}\hfill
  \subfloat{\includegraphics[width=9mm]{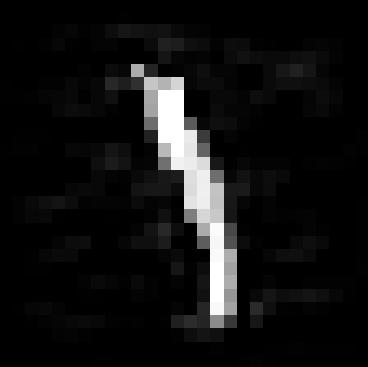}\label{fig:f1}}\hfill
  \subfloat{\includegraphics[width=9mm]{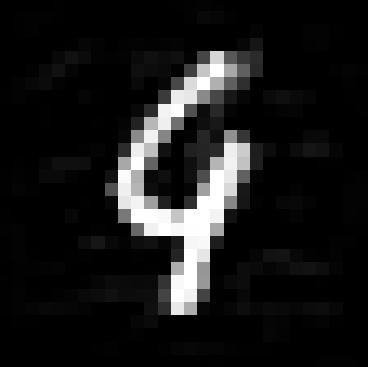}\label{fig:f2}}\hfill
  \subfloat{\includegraphics[width=9mm]{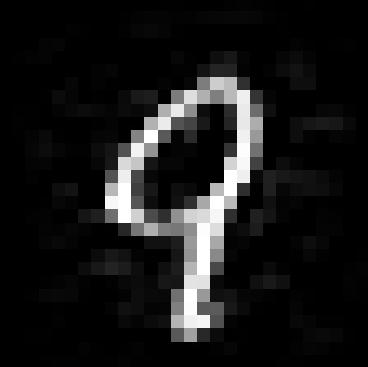}\label{fig:f2}}\hfill
  \subfloat{\includegraphics[width=9mm]{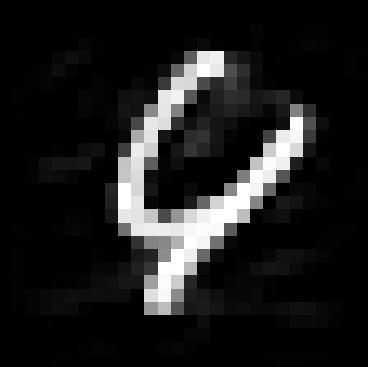}\label{fig:f2}}\hfill
  \caption{MNIST-Digit images attacked with the FGS algorithm. The upper row has unperturbed images. The middle row has images perturbed using gradients of an FC classifier with $l_2$ norm 2.5. The bottom row has images perturbed using gradients of a CNN classifier with $l_2$ norm 1.5.}
  \label{fig:attacked_imgs_mnist}
\end{figure}

\begin{figure}[!tbp]
  \centering
  \subfloat{\includegraphics[width=9mm]{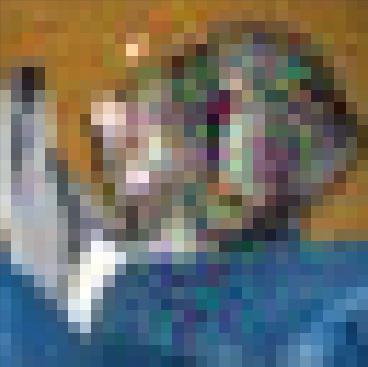}\label{fig:f1}}\hfill
  \subfloat{\includegraphics[width=9mm]{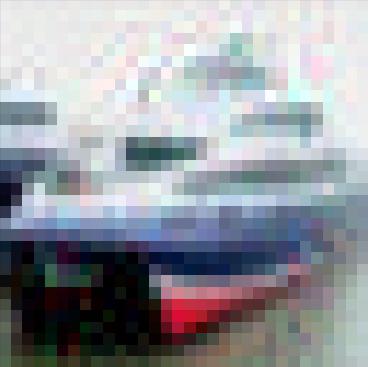}\label{fig:f1}}\hfill
  \subfloat{\includegraphics[width=9mm]{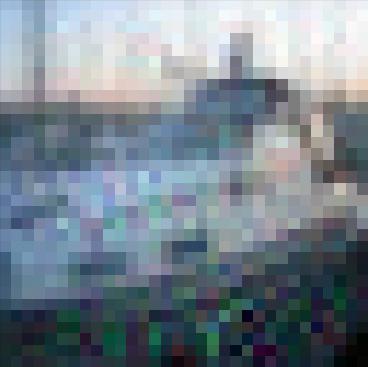}\label{fig:f1}}\hfill
  \subfloat{\includegraphics[width=9mm]{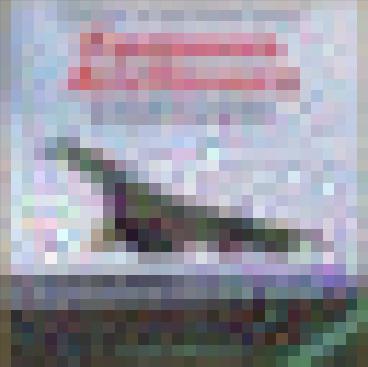}\label{fig:f1}}\hfill
  \subfloat{\includegraphics[width=9mm]{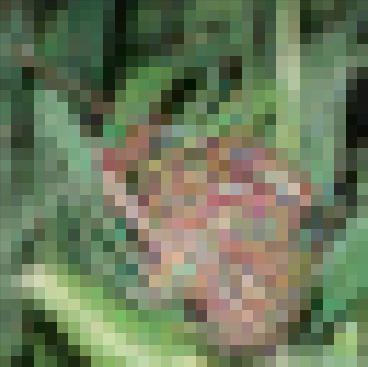}\label{fig:f1}}\hfill
  \subfloat{\includegraphics[width=9mm]{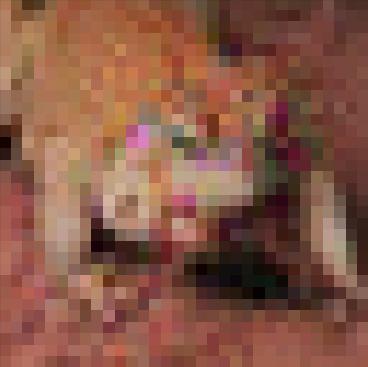}\label{fig:f1}}\hfill
  \subfloat{\includegraphics[width=9mm]{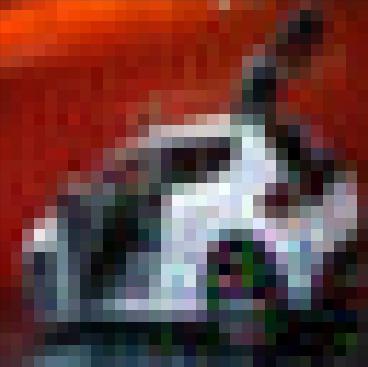}\label{fig:f2}}\hfill
  \subfloat{\includegraphics[width=9mm]{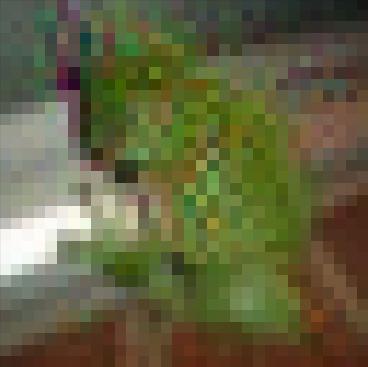}\label{fig:f2}}\hfill
  \subfloat{\includegraphics[width=9mm]{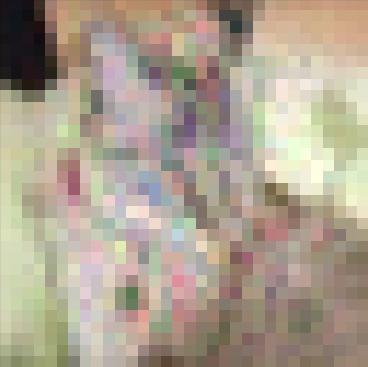}\label{fig:f2}}\hfill
  \subfloat{\includegraphics[width=9mm]{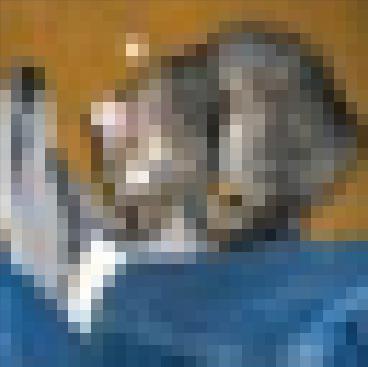}\label{fig:f1}}\hfill
  \subfloat{\includegraphics[width=9mm]{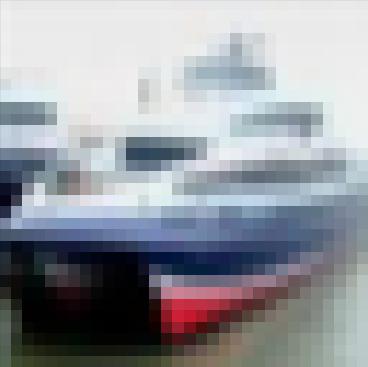}\label{fig:f1}}\hfill
  \subfloat{\includegraphics[width=9mm]{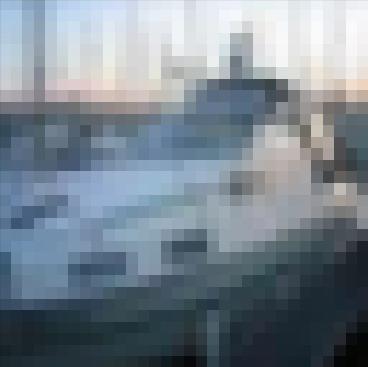}\label{fig:f1}}\hfill
  \subfloat{\includegraphics[width=9mm]{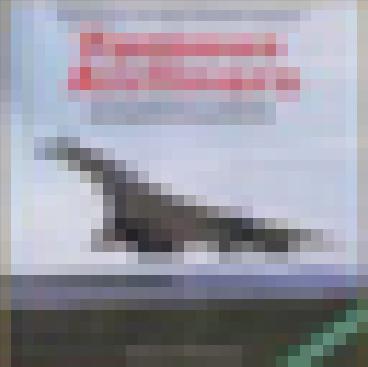}\label{fig:f1}}\hfill
  \subfloat{\includegraphics[width=9mm]{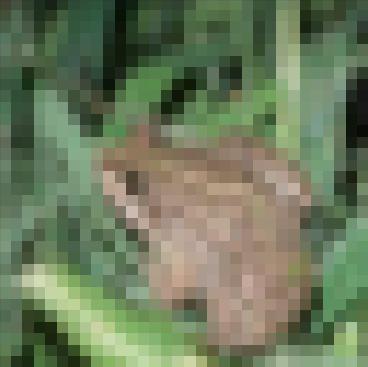}\label{fig:f1}}\hfill
  \subfloat{\includegraphics[width=9mm]{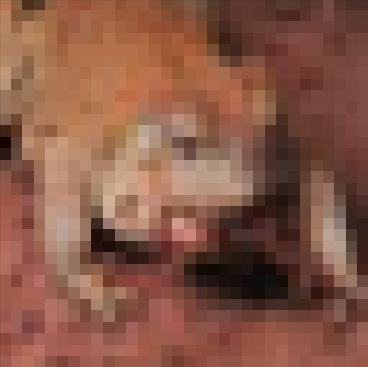}\label{fig:f1}}\hfill
  \subfloat{\includegraphics[width=9mm]{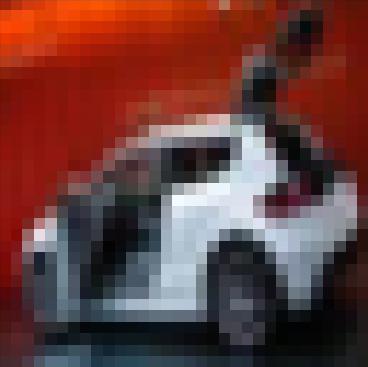}\label{fig:f2}}\hfill
  \subfloat{\includegraphics[width=9mm]{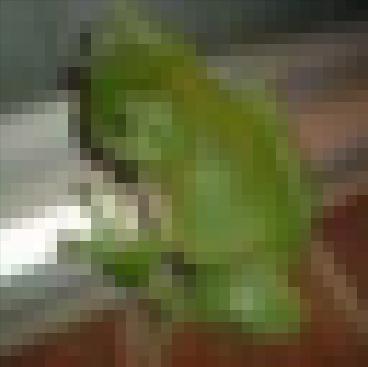}\label{fig:f2}}\hfill
  \subfloat{\includegraphics[width=9mm]{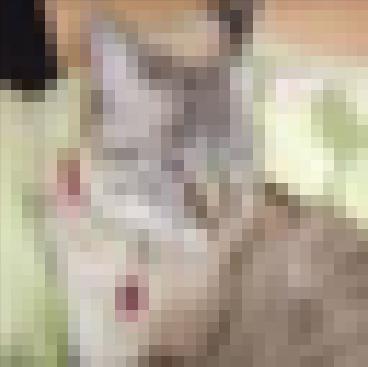}\label{fig:f2}}\hfill
  \caption{Images from the CIFAR-10 dataset attacked with the FGS algorithm with an $l_2$ norm of 1.7. The upper row has unperturbed images. The lower row has perturbed images.}
  \label{fig:attacked_imgs_cifar}
\end{figure}

We evaluate the defense sensitivity to the choice of attacked data used for training by varying the architecture and attack algorithm. To train the DAE with data perturbed according to the FGS attack, the $l_2$ norm was chosen such that the \textbf{perturbation is simultaneously effective and reasonably imperceptible} in the attacked images. For the MNIST-Digit dataset, this was achieved by using an $l_2$-norm of 1.5 when using gradients from a CNN, and an $l_2$-norm of 2.5 when using gradients from an FC network, as shown in Figure \ref{fig:attacked_imgs_mnist}. For the CIFAR-10 dataset, an $l_2$ norm of 1.7 delivered reasonably imperceptibly attacked images, as shown in Figure \ref{fig:attacked_imgs_cifar}.
We say that the defense is \textit{$<${}architecture-type$>${}-trained} or \textit{$<${}attack$>${}-trained} to refer to the architecture type or attack algorithm used to simulate the perturbations, respectively. Further, we say that the defense is \textit{ensemble-architecture-type-trained} or \textit{ensemble-attack-trained} when both FC and CNN architectures or when all three attack algorithms are used to simulate the perturbations, respectively\footnote{ Code available at https://codeocean.com/capsule/7339381/tree/v1}.
\subsubsection{Varying the Architecture Type}
Here, we vary the architecture type used to train the defense for a particular attack type. 
For each attack, we trained two defenses using gradients of one architecture type, and one defense trained using gradients of an ensemble of both architecture types. We only performed this with the MNIST-Digit dataset, since the architecture type cannot be varied with the CIFAR-10 dataset.

\subsubsection{Varying the Attack Algorithm}

Next, we trained the defense with data attacked using gradients of a particular architecture type while varying the attack algorithm used. For the MNIST-Digit dataset, we used an FC architecture.With the CIFAR-10 dataset, we took a step further and trained the defense with data that is attacked using gradients from an ensemble of classifiers with varying CNN architectures.
\section{Results}

\subsubsection{Performance Metric}
We first measure the \textit{accuracy improvement} as the pre-defense accuracy subtracted from the post-defense accuracy. We then compute the percent increase in accuracy improvement when using the proposed defense from that when using traditional defenses:
\begin{equation}
    \% \text{increase in accuracy improvement} = \frac{p-t}{abs(t)}*100,
\end{equation}
where $p$ is the accuracy improvement when using our proposed defense,
and $t$ is the average accuracy improvement when using all traditional defenses for which the attacker's choice of architecture type (or attack) was not involved.
When we vary the architecture type, the traditional defense considered is the defense trained with the same attack as the attacker's, based on an architecture type different from the attacker's. In the absence of an attack while varying the architecture type, the traditional defenses considered are the two defenses trained with a particular attack based on a single architecture type.
When we vary the attack algorithm, the traditional defenses considered are the two defenses trained with an attack algorithm different from the attacker's, and the one defense trained using the two attack algorithms that the attacker did not use. In the absence of an attack while varying the attack algorithm, the traditional defenses considered are the three defenses trained with one attack each, and the three defenses trained with an ensemble of two attacks.

\subsubsection{Varying the Architecture Type}

\begin{figure}
\centerline{\includegraphics[width=90mm]{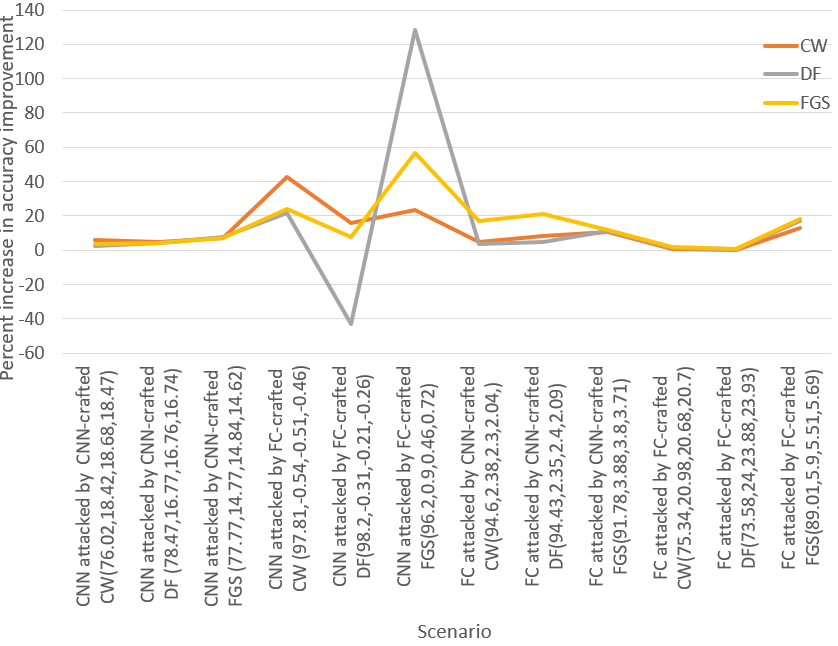}}
\caption{Percent increase in accuracy improvements for the MNIST-Digit dataset when using the ensemble-architecture-type-trained defense. The numbers in parentheses after each scenario description indicate the following: post-attack accuracy before defense, average accuracy improvement when using the CW, DF and FGS traditional defenses respectively.}
\label{fig:vary_arch}
\end{figure}
In most cases, our ensemble-architecture-type-trained defense improves the accuracy, and often significantly, as shown through the positive percent increases in Figure \ref{fig:vary_arch}. The only case with a negative percent increase corresponds to a very weak attack that leaves the classifier achieving $98.2\%$ without defense. We also note that in case of such a weak attack, using the considered single-architecture-type-trained-defenses also worsens the accuracy compared to the no-defense case. Hence it may be smarter to not use a preprocessing defense when the attack is so weak.



With no attack, our ensemble-architecture-type-trained defense decreases the accuracy improvement by not more than 10.68\%, and increases it by up to 37.7\%. 

\subsubsection{Varying the Attack}

\begin{figure}[!tbp]
  \centering
  \subfloat[MNIST-Digit dataset.]{\includegraphics[width=40mm]{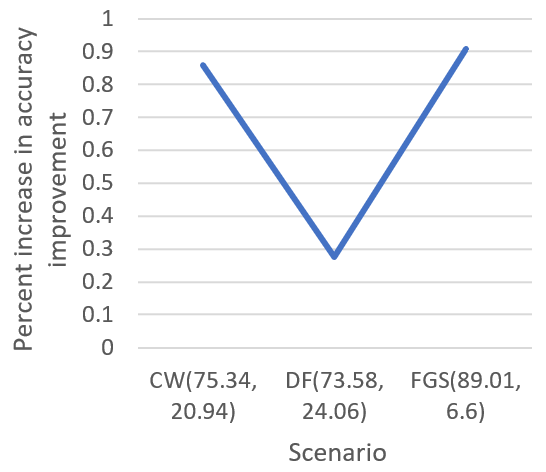}\label{fig:f1}}
  \hfill
  \subfloat[CIFAR-10 dataset.]{\includegraphics[width=40mm]{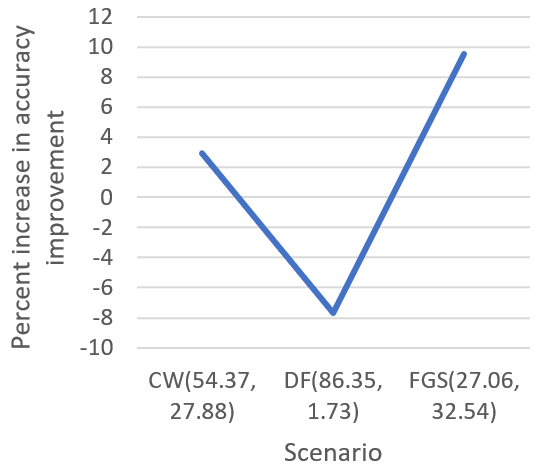}\label{fig:f2}}
  \caption{Percent increases in accuracy improvements when using an ensemble-attack-trained defense. The numbers in parentheses after each scenario description indicate the following: post-attack accuracy before defense, average accuracy improvement when using traditional defenses.}
  \label{fig:vary_attack}
\end{figure}

Using our ensemble-attack-trained defense generally yields an increase in accuracy improvement compared to using the traditional defenses, as shown in Figure \ref{fig:vary_attack}. The only instance when this accuracy improvement is negative is when a DeepFool attack is used against the CIFAR-10 classification task. As earlier, this negative accuracy improvement happens only when the attack is weak.

With no attack, our ensemble-attack-trained defense changes the accuracy improvement by +43.86\% and -10.12\% for the MNIST-Digit and CIFAR-10 datasets, respectively.

\section{Discussion}
In our experiments, training the defense with an ensemble of architecture types made a larger impact than training the defense with an ensemble of attacks. We believe that this is because while generating attacks using different algorithms, we adjusted the hyperparameters such that the levels of perceptibility of the attacked images are similar. In conclusion, if the attacker is constrained to limit the perceptibility of the attack, then the choice of attack in training the defense does not make a significant impact, as evident from Figure \ref{fig:vary_attack}.

We also observe that it is crucial for the attacker to know the architecture type of the victim classifier, as illustrated by the pre-defense accuracies in Figure \ref{fig:vary_arch}. From the defender's perspective, if the attack is weak, it is in their best interest to avoid using a pre-processing defense. This makes it important to develop detection strategies to determine how strong an attack is. During test-time, if the attack is found to be weaker than a certain threshold, using a pre-processing defense should be avoided to avoid a decrease in accuracy. Other problems that we plan to consider for future work include the case when the test noise model cannot be included in training (see e.g., \cite{sigproc2}), creating adaptive defenses by classifying noise types (see e.g., \cite{sigproc6}) and creating a noise dictionary (see e.g., \cite{sigproc9}), as well as incorporating uncertainty prediction in the defender's model (see e.g., \cite{sigproc15}).


\bibliographystyle{IEEEtran}

\appendix
\section*{Details of Attack Generation}

All attacks were generated using the TensorFlow Cleverhans library and were untargeted attacks.
For the MNIST-Digit dataset, the CW attack was generated with 4 binary search steps, a maximum of 60 iterations, a learning rate of 0.1, a batch size of 10, an initial constant of 1.0, and the abort\_early parameter was set to True.
For the CIFAR-10 dataset, the CW attack was generated with 6 binary search steps, a maximum of 10000 iterations, a learning rate of 0.7, a batch size of 25, an initial constant of 0.001, and the abort\_early parameter was set to True.

\section*{Architecture of Denoising Autoencoders}

\setcounter{table}{1}

The MNIST DAE has architecture FC-784-256-128-81-128-256-784. There is no activation in any layer except the last layer, which has sigmoid activation. The architecture of the CIFAR-10 DAE is shown in Table 1 
 \begin{table}[h]
 \renewcommand\thetable{I} 
 \begin{tabular}{ p{8cm} }
 \hline
 Conv 3x3x64, ReLU\\
 Conv 3x3x32, ReLU\\
 Max Pool 2x2\\
 Conv 3x3x3, ReLU\\
 Conv 3x3x32, ReLU\\
 Upsampling 2x2\\
 Conv 3x3x64, ReLU\\
 Conv 3x3x64, Sigmoid\\
 \hline
 \end{tabular}
  \label{table: cifar10_dae_arch}
 \caption{CIFAR-10 DAE architecture}
 \end{table}
\section*{Training Details of all Neural Networks}

Table 2 shows the training details of all neural networks used in this letter. The default values were used for all hyperparameters not shown in the table.

\begin{table}[h]
\renewcommand\thetable{II} 
\centering
\begin{tabular}{|c|c|c|c|c|c|}
\hline
Parameters & MNIST FC & MNIST CNN & MNIST DAE & CIFAR CNN & CIFAR DAE\\
 \hline
 Loss & Categorical Crossentropy & Categorical Crossentropy & MSE & Categorical Crossentropy & MSE\\
 Optimizer & Adam & Adam & Adam & RMSprop & Adam\\
 Learning Rate & 0.001 & 0.001 & 0.001 & 0.001 & 0.001\\
 Batch Size & 200 & 200  & 200 & 64 & 256\\
 Epochs & 100 & 20 & 150 & 150 & 150\\
 \hline
\end{tabular}
\label{table: all_params}
\caption{Training Parameters for the Neural Networks}
\end{table}

\end{document}